\documentclass[10pt,twocolumn,letterpaper]{article}

\usepackage{iccv}              
\usepackage{multirow}
\usepackage{array} 
\usepackage{makecell}
\usepackage{float}

\definecolor{iccvblue}{rgb}{0.21,0.49,0.74}
\usepackage[pagebackref,breaklinks,colorlinks,allcolors=iccvblue]{hyperref}

\def\paperID{5309} 
\def\confName{ICCV}
\def\confYear{2025}

\title{ MCOP: Multi-UAV Collaborative Occupancy Prediction}

\author{
  \textbf{Zefu Lin}$^{1,2,3}$\quad
  \textbf{Wenbo Chen}$^{2,3}$\quad
  \textbf{Xiaojuan Jin}$^{2,3}$\quad 
  \textbf{Yuran Yang}$^{5,6}$ \quad
  \textbf{Lue Fan}$^{2,3}$ \quad \\
  \textbf{Yixin Zhang}$^{6}$ \quad
  \textbf{Yufeng Zhang}$^{1}$ \quad
  \textbf{Zhaoxiang Zhang}$^{1,2,3,4}$ \thanks{Corresponding author}
  \\
$^1$ University of Chinese Academy of Sciences (UCAS)\\
$^2$ Institute of Automation, Chinese Academy of Sciences (CASIA)\\
$^3$ New Laboratory of Pattern Recognition (NLPR)\\
$^4$ State Key Laboratory of Multimodal Artificial Intelligence Systems (MAIS)\\
$^5$ Beijing University of Posts and Telecommunications (BUPT)  
$^6$ Tencent \\
}

\begin{document}
\maketitle
\begin{abstract}
Unmanned Aerial Vehicle (UAV) swarm systems necessitate efficient collaborative perception mechanisms for diverse operational scenarios. 
Current Bird's Eye View (BEV)-based approaches exhibit two main limitations: bounding-box representations fail to capture complete semantic and geometric information of the scene, and their performance significantly degrades when encountering undefined or occluded objects.
To address these limitations, we propose a novel multi-UAV collaborative occupancy prediction framework. Our framework effectively preserves 3D spatial structures and semantics through integrating a Spatial-Aware Feature Encoder and Cross-Agent Feature Integration. To enhance efficiency, we further introduce Altitude-Aware Feature Reduction to compactly represent scene information, along with a Dual-Mask Perceptual Guidance mechanism to adaptively select features and reduce communication overhead.
Due to the absence of suitable benchmark datasets, we extend three datasets for evaluation: two virtual datasets (Air-to-Pred-Occ and UAV3D-Occ) and one real-world dataset (GauUScene-Occ). 
Experiments results demonstrate that our method achieves state-of-the-art accuracy, significantly outperforming existing collaborative methods while reducing communication overhead to only a fraction of previous approaches. 
\end{abstract}

\section{Introduction}
\label{sec:Introduction}

\begin{figure}[t]
  \centering
    \includegraphics[width=1.0\linewidth]{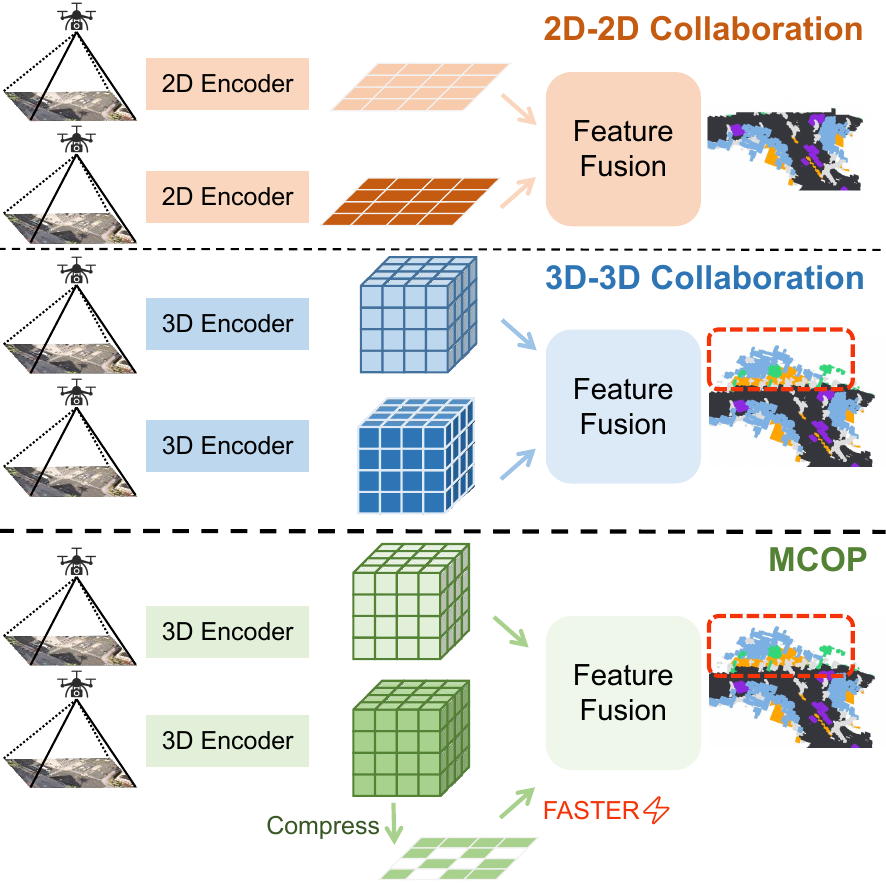}
    \caption{\textbf{Comparison of Multi-UAV collaborative occupancy prediction and other collaborative methods.} 2D-to-2D method lacks height information and cannot effectively reconstruct the 3D details of the scene. 3D-to-3D method requires transmitting high-dimensional occupancy features, which demands high bandwidth and affects real-time performance. MCOP compresses occupancy features, balancing transmission rate and prediction quality.
}
  \label{fig:show}
  \vspace{-12pt}
\end{figure}

Unmanned Aerial Vehicles (UAVs) are increasingly used in applications such as smart cities~\cite{smartcity}, traffic management, and emergency response~\cite{fan2023few}.
These applications require advanced environmental perception capabilities that single-UAV systems inherently lack due to their limited field of view and susceptibility to occlusions.
To address these challenges, multi-UAV collaborative perception has emerged as a promising solution by integrating observations from multiple viewpoints to enhance scene understanding.

Current multi-UAV perception systems typically project image features into a unified Bird's-Eye-View (BEV) coordinate system for 3D object detection~\cite{zhang2019eye, wu2024cmp, wang2024deepaccident, cui2022coopernaut}. 
While effective for identifying specific objects, these approaches fail to capture the rich geometric and semantic details of the environment, such as irregularly shaped obstacles or partially occluded structures. 
Furthermore, BEV-based features lack altitude information, which is particularly crucial for UAVs due to their reliance on 3D spatial awareness.
To overcome these limitations, recent research has explored the use of 3D occupancy prediction~\cite{occ3d}, which represents the environment as a voxel grid encoding both occupancy status and semantic categories. 
Unlike bounding box-based methods, occupancy prediction provides a comprehensive understanding of the 3D scene, including free space, occupied regions, and undefined obstacles.
However, extending occupancy prediction to multi-UAV collaborative scenarios introduces a new set of challenges.
UAVs typically operate at altitudes $10 \times$ higher than ground-based autonomous vehicles, requiring them to perceive a broader range of scenes—from ground surfaces to buildings and aerial objects. 
This significantly expands the feature space for occupancy representation, making real-time processing and communication infeasible with existing methods~\cite{cohff}.

In this paper, we propose Multi-UAV Collaborative Occupancy Prediction (MCOP), a vision-centric framework designed to address the unique challenges of UAV-based 3D scene understanding. 
Our approach leverages the rich geometric and semantic information provided by occupancy prediction while overcoming the computational and communication bottlenecks associated with multi-UAV collaboration.
The core of MCOP lies in its novel visual feature representation and fusion mechanisms, which are specifically designed for UAV viewpoints. 
First, we introduce the Spatial-Aware Feature Encoder, which transforms RGB images into 3D occupancy features using a combination of Voxel-Image Attention and Cross-Voxel Attention. 
This encoder effectively captures detailed scene geometry and semantics without relying on depth sensors, making it suitable for resource-constrained UAV platforms. 
To address the high-dimensionality of occupancy features, we propose Altitude-Aware Reduction, a compression mechanism that retains critical height information while reducing feature dimensions. 
This is achieved by encoding vertical pillars into 2D BEV representations, significantly reducing communication overhead without sacrificing perceptual accuracy.
Furthermore, we develop Dual-Mask Perceptual Guidance, a dynamic feature selection mechanism that identifies and transmits only the most relevant visual information across UAVs. 
By leveraging support masks (high-confidence regions) and request masks (low-confidence regions), this module minimizes redundant data transmission while ensuring robust perception in occluded or complex scenes.
Finally, the Cross-Agent Feature Integration module fuses local and received features into a unified 3D occupancy representation, enabling comprehensive scene understanding across multiple UAVs.

Because 3D occupancy labeling is expensive, no public dataset currently supports multi-UAV collaborative semantic occupancy prediction. To address this gap, we extend three datasets for our evaluation: two CARLA-based virtual datasets, Air-to-Pred-Occ~\cite{Wang2024DronesHD} and UAV3D-Occ~\cite{uav3d2024}, and one real-world dataset, GauUScene-Occ~\cite{xiong2024gauuscene}.
We enrich each with 3D occupancy annotations, thereby filling a crucial gap in UAV collaborative perception research. 
Inspired by \cite{occ3d}, we employ a streamlined method to derive suitable occupancy ground truth for these aerial scenarios.

Experimental results demonstrate that collaborative perception consistently outperforms single-UAV perception in semantic occupancy prediction, benefiting from enhanced spatial coverage and information sharing. 
Comparative analysis with adapted autonomous driving approaches BEVDet~\cite{bevdet} and PanoOcc~\cite{panoOcc} shows that our method achieves higher mIoU on all evaluated datasets with significantly reduced communication overhead (0.23 MB vs. 17.50 MB and 19.14 MB, respectively).

\noindent{\textbf{Contributions}} Our key contributions are:
\begin{itemize}
    \item \textbf{Occupancy-Based Multi-UAV Perception Framework.} To our knowledge, we propose the first collaborative occupancy prediction framework for multi-UAV systems. Our method addresses key limitations of BEV-based approaches by effectively preserving rich semantic and geometric information including occluded objects.
    \item \textbf{High-Efficiency Collaboration Strategy.} Altitude-Aware Reduction and Dual-Mask Perceptual Guidance significantly lower communication overhead while preserving essential 3D features, thus supporting real-time collaboration among UAVs.
    \item \textbf{Enriched Collaborative Datasets.} We extend three datasets with occupancy annotations, offering a new benchmark for multi-UAV semantic occupancy prediction and fostering further exploration in aerial 3D perception research.
\end{itemize}

\section{Related Work}
\label{sec:related work}

\subsection{Collaborative Prediction}  
In multi-agent systems, sharing information across perception nodes (vehicles, infrastructure, etc.) effectively expands a node’s field of view and mitigates occlusion-induced degradation~\cite{smartcity,qiao2023adaptive}. In large-scale scenarios, collaborative perception significantly improves detection accuracy and robustness over individual perception~\cite{han2023collaborative}.

Collaboration strategies are typically categorized as early, intermediate, or late, based on the fusion stage of sensing modalities~\cite{Wang2024DronesHD}. Early collaboration fuses raw data at the input layer~\cite{arnold2020cooperative}, maximizing shared content but requiring high bandwidth. Late collaboration merges target predictions at the output~\cite{cui2022coopernaut}, conserving bandwidth but often amplifying accumulated noise. Intermediate collaboration, focusing on feature-level fusion~\cite{vadivelu2021learning}, achieves a balanced trade-off between communication cost and accuracy~\cite{uavdetectionsurvey}.

Information-sharing strategies differ among nodes. Some methods share all data to maximize coverage, at the cost of bandwidth. To reduce redundancy, dynamic communication strategies like Who2com~\cite{liu2020who2com} and When2com~\cite{when2com} use attention or scheduling to determine optimal communication timing and partners. Where2comm~\cite{where2comm} further selects informative local features based on regional uncertainty.

Feature fusion began with simple operations~\cite{zhang2024vision}. F-Cooper~\cite{chen2019f} uses element-wise max for voxel-level fusion; CoHFF~\cite{cohff} incorporates similarity-based weighting to exploit complementary, low-confidence features. V2VNet~\cite{wang2020v2vnet} applies a variational graph network, while DiscoNet~\cite{li2021learning} introduces matrix-valued weights for fine-grained attention. Recent transformer-based models such as V2X-ViT~\cite{xu2022v2x} and CoBEVT~\cite{xu2022cobevt} use multi-agent attention for multi-camera fusion. CoCa3D~\cite{hu2023collaboration} enhances depth prediction using uncertainty to improve cross-view fusion. However, these methods mainly target 2D feature fusion; moving to 3D requires additional mechanisms to preserve real-time performance.

\subsection{Occupancy Prediction}
Unlike detection-based methods, occupancy prediction estimates the semantic state of each voxel. Reconstructing 3D scenes from visual input demands complete geometry and semantic reasoning, posing challenges due to high dimensionality and data sparsity~\cite{wei2023surroundocc,pan2024renderocc,cao2022monoscene,flashocc,zhang2024vision}.


To mitigate 2D-to-3D projection uncertainty, FB-BEV~\cite{li2023fb} uses both forward and backward projections and applies depth-consistency weighting. 
Addressing height-information loss in standard BEV projection, the TPV family~\cite{huang2023tri,silva2024s2tpvformer} exploits three complementary viewpoints (top, front, side).
Alternatively, some methods directly process 3D features. 
MiLO~\cite{myeongjin2023milo} uses 3D ResNet~\cite{he2016identity} and FPN~\cite{lin2017feature}, PanoOcc~\cite{panoOcc} merges spatiotemporal voxel queries for detailed 3D information. 
Voxformer~\cite{li2023voxformer} employs sparse voxel queries to index 2D features via camera projection.
COTR~\cite{cotr} leverages geometry priors and explicit–implicit transforms to reduce voxel sparsity.

The primary challenge lies in effectively learning high-dimensional and sparse 3D features. Methods based on BEV, TPV, or direct 3D operations each address the core issue of representation sparsity and depth inference from a different angle. This becomes especially difficult when aligning multi-view 2D inputs with 3D space in large-scale or dynamic settings. In multi-UAV cooperative perception, for example, frequent viewpoint shifts, larger feature dimensions, and greater motion variability make stable feature extraction and alignment even more demanding. Nonetheless, coupling occupancy prediction with collaborative strategies—such as leveraging uncertainty or visibility masks to reduce redundant transmissions—can still deliver a refined, complete representation of the environment.
\section{Methodology}
\label{sec:Methodology}
\begin{figure*} 
    \centering 
    \includegraphics[width=1.0\linewidth]{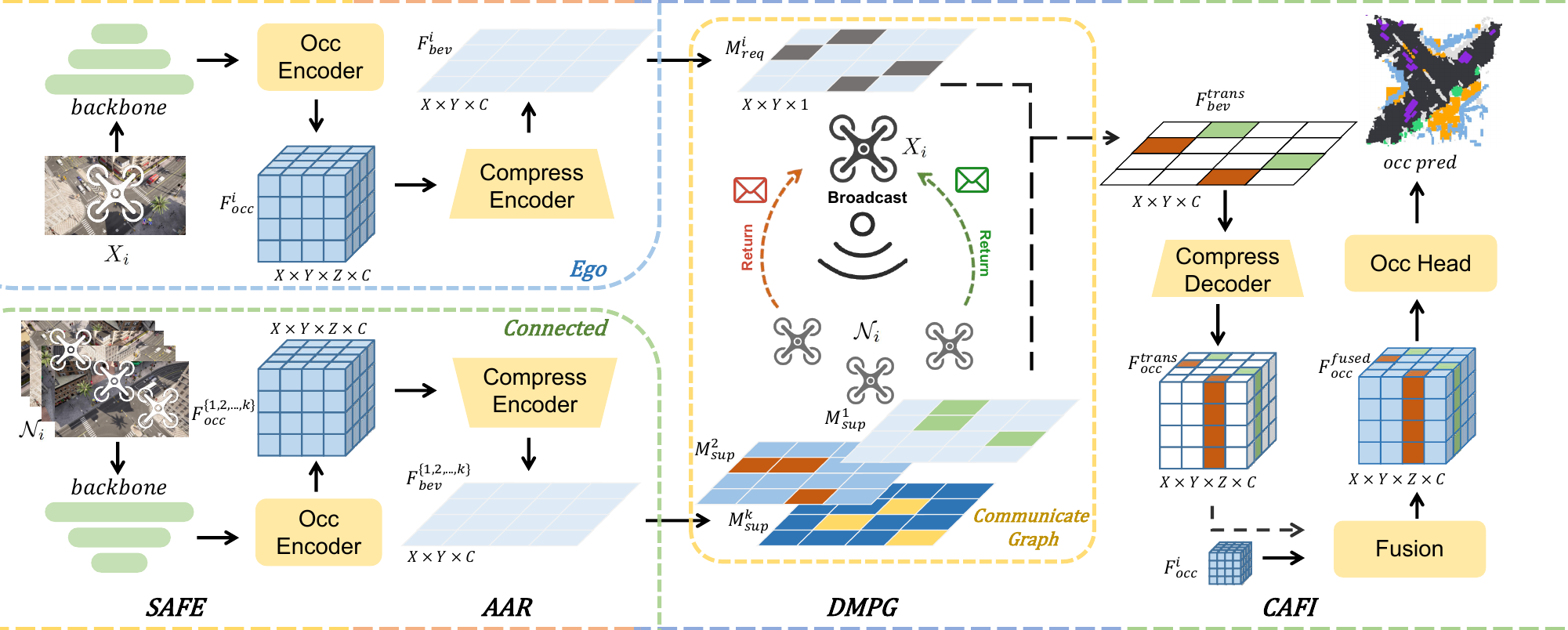} 
    \caption{\textbf{The overall framework of MCOP.} Each UAV uses an image backbone to extract multi-scale features, which are processed by the \emph{Spatial-Aware Feature Encoder \bf{(SAFE)}} to generate 3D occupancy features. The \emph{Altitude-Aware Reduction \bf{(AAR)}} compresses these 3D features into compact 2D BEV representation for efficient communication. \emph{Dual-Mask Perceptual Guidance \bf{(DMPG)}} coordinates the sharing of relevant information among UAVs based on perception quality. 
    The \emph{Cross-Agent Feature Integration \bf{(CAFI)}} module fuses local and received features into unified 3D representation. 
    Finally, the Occ Head predicts 3D occupancy segmentation, resulting in comprehensive environmental perception among UAVs.} 
    \vspace{-10pt} 
    \label{fig:framework} 
\end{figure*}

Our MCOP framework consists of four key modules, namely Spatial-Aware Feature Encoder, Altitude-Aware Reduction, Dual-Mask Perceptual Guidance and Cross-Agent Feature Integration. It achieves efficient inter-UAV collaborative prediction with minimal accuracy cost by transmitting encoded spatial-aware occupancy features.
\subsection{Problem Setup}
In the multi-UAV collaborative 3D occupancy prediction task, we define the UAV network by a global communication network, represented as an undirected graph $\mathcal{G} = (\mathcal{X}, \mathcal{L})$, where $\mathcal{X} = \left\{X_1, X_2, \dots, X_n\right\}$ denotes all UAVs, and $\mathcal{L} = \left\{L_{ij} \mid i, j \in [1, n]\right\}$ where ${L}_{ij}$ denotes the communication links between UAV $X_i$ and $X_j$.
For each UAV $X_i$, the set of connected UAVs is represented as $\mathcal{N}_i = \left\{ {X}_j \mid {L}_{ij} \in \mathcal{L}, j \in [1, k]\right\}$, where $\mathcal{N}_i$ denotes all the UAVs directly communicating with UAV ${X}_i$.
UAV $X_i$ takes RGB images $\mathcal{I}_i \in \mathbb{R}^{H \times W \times 3}$ as input, and outputs 3D occupancy prediction $\mathcal{O} \in \mathbb{R}^{ X \times Y \times Z }$ with certain semantic categories, and $X$, $Y$, $Z$ are dimensions of 3D occupancy voxel space.
Inspired by~\cite{cohff}, the optimization problem is defined as follows
\begin{multline}
    \max_{\theta, {F}} \sum_{X_i \in \mathcal{X}} g\big(\Phi_\theta (\mathcal{I}_i, \{ {F}_{j \rightarrow i} \mid X_j \in \mathcal{N}_i \}), \mathcal{O}^{gt}_{i}\big), \\
    \text{s.t.} \quad \sum_{X_i \in \mathcal{X}} \sum_{X_j \in \mathcal{N}_i} \lvert {F}_{j \rightarrow i} \rvert \leq B ,
\end{multline}
where $\Phi_\theta$ represents the model parameterized by $\theta$, ${F}_{j \rightarrow i}$ denotes the features transmitted from UAV ${X}_j$ to UAV ${X}_i$, and $g(\cdot, \cdot)$ represents the function to evaluate the predicted occupancy against the ground truth $\mathcal{O}^{gt}_{i}$. $B$ denotes the dynamic communication volume constraint, which may vary depending on hardware conditions. The optimization objective is to maximize the overall perception effectiveness within the communication upper bound $B\in \mathbb{R}^+$. 

\subsection{Overall Architecture}
This section introduces the overall architecture of MCOP.
As illustrated in Fig.~\ref{fig:framework}, each UAV independently takes RGB images as input and uses Spatial-Aware Feature Encoder~\ref{sec:safe}, which includes an image backbone and an Occupancy Encoder, to generate 3D occupancy features. 
To minimize the data required for collaborative communication, these 3D occupancy features are compressed into 2D BEV features via the Altitude-Aware Reduction~\ref{sec:aar}.
It encodes features based on effective spatial information, thus reducing bandwidth requirements. 
Up to this stage, each UAV operates independently without interaction.

In the Dual-Mask Perceptual Guidance module~\ref{sec:dmpg}, each UAV assesses the perceptual quality of its local regions based on its 2D BEV features. 
It then generates two types of masks: a support mask, representing regions with high perceptual confidence, and a request mask, indicating areas with low perceptual confidence that require assistance from other UAVs.
During each communication phase, the ego UAV broadcasts its request mask to solicit assistance from other connected UAVs. 
Connected UAVs project their support masks into the ego UAV's perception space and compute the intersection with the ego's request mask to determine the regions requiring collaboration. 
The connected UAVs then transmit the corresponding compressed feature data for these regions to the ego UAV, enabling collaborative perception.
This interaction ensures that the ego UAV receives only the necessary, high-confidence information.

After receiving information from other UAVs, the ego UAV applies Cross-Agent Feature Integration~\ref{sec:cafi} to combine its 3D occupancy features with the received 2D features, yielding 3D fused occupancy representation. 
This fused representation is subsequently used by the task processing head for 3D occupancy segmentation. 
These modules together enable our method to achieve efficient collaborative perception among multiple UAVs. The following sections describe each module in detail.
In the following sections, we provide detailed descriptions of each module.

\subsection{Spatial-Aware Feature Encoder \label{sec:safe}}
For input RGB images, we first use pretrained backbone network (e.g. ResNet~\cite{he2016identity}) to extract image features. 
To capture detailed scene information without relying on depth, we follow ~\cite{panoOcc} and define a set of 3D voxel queries $\mathbf{Q} \in \mathbb{R}^{X \times Y \times Z \times C}$, where $C$ represents the feature channels, and $X$, $Y$, $Z$ are the voxel grid dimensions.
We propose Voxel-Image Attention to bridge feature extraction and voxel representation, which uses deformable attention~\cite{zhu2021deformable} to associate each voxel query $\mathbf{q}$ at $(x, y, z)$ with relevant image features 
This Voxel-Image Attention (VIA) can be defined as
\begin{equation}
    \text{VIA}(\mathbf{q}, f(\mathcal{I}_i)) = \sum_{\eta=1}^{\mathcal{P}_s} \text{DA} (\mathbf{q}, \delta(\mathbf{Ref}_{(x, y, z)}^{\eta}),  f(\mathcal{I}_i)), 
\end{equation}
where $f$ is image backbone, ${\mathcal{P}_s}$ is the number of sampling points per voxel query, and $\delta(\mathbf{Ref}_{(x, y, z}^{\eta})$ denotes the $\eta$-th sampling point projected onto the voxel grid using projection matrix $\delta$. $\text{DA}$ denotes deformable attention.
During this process, the querying paradigm~\cite{bevformer} efficiently transforms perspective view features into voxel space representations, reducing computational complexity.
We next apply Cross-Voxel Attention (CVA) to establish connections between voxel queries, which is defined as 
\begin{equation}
    \text{CVA}(\mathbf{q}, \mathbf{Q}) = \sum_{\eta=1}^{\mathcal{P}_r} \text{DA} (\mathbf{q}, \mathbf{Ref}_{(x, y, z)}^{\eta}, \mathbf{Q}),
\end{equation}
where ${\mathcal{P}_r}$ is the number of reference points per voxel query. 

These operations allow voxel queries to interact both with image pixels and with each other, enriching the geometric and semantic content. 
Finally, we obtain the occupancy feature ${F}^{i}_{\mathrm{occ}} \in \mathbb{R}^{X \times Y \times Z \times C}$.

\begin{figure}[t]
  \centering
    \includegraphics[width=1.0\linewidth]{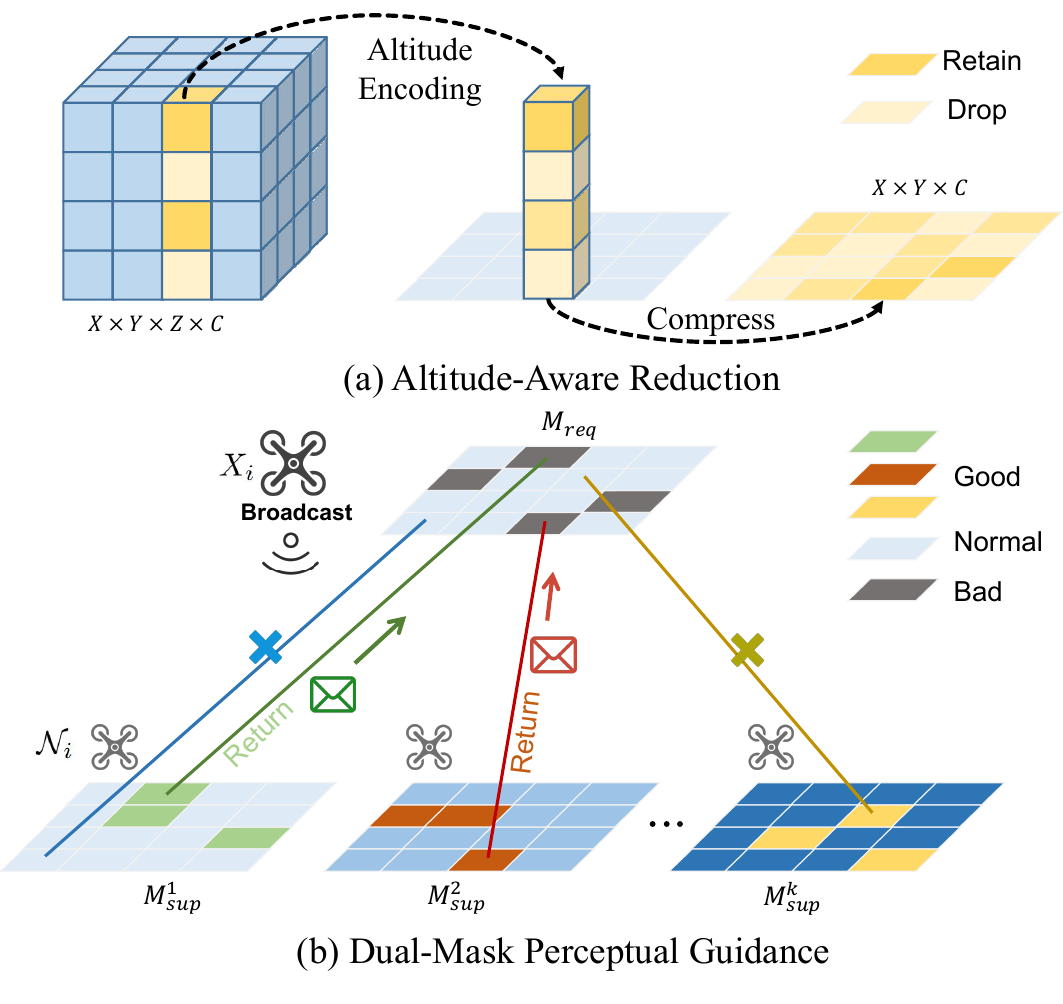}
    \caption{
    \textbf{Illustration of Altitude-Aware Reduction and Dual-Mask Perceptual Guidance.}
     (a) demonstrates how a pillar in the 3D occupancy feature is compressed into a grid in the 2D BEV feature. 
     (b) illustrates the details of DMPG, where the ego UAV first broadcasts its request mask within the network, and then the connected UAVs select and transmit high-quality features that are needed by the ego UAV.}
    
  \label{fig:detail}
  \vspace{-12pt}
\end{figure}

\subsection{Altitude-Aware Reduction \label{sec:aar}} 
In multi-UAV perception system, common collaboration strategy is to transmit encoded image features to share environmental information. 
This approach is more efficient than transmitting raw images and outperforms sharing post-processed perception results.
However, for occupancy prediction, the encoder output is 3D features, and transmitting 3D features directly still requires significant bandwidth.
Given that only about 5\% of the spatial areas are occupied, we designed Altitude-Aware Reduction(AAR) to compress 3D occupancy features into 2D BEV features, thereby reducing communication costs.

As shown in Fig.~\ref{fig:detail}(a), we first normalize the 3D features using sigmoid function and apply a threshold $\theta$ to filter valid spatial points. 
For each pillar, we create an altitude embedding along the Z-axis as $[1, 2, 3, \ldots, Z-1]$ and a binary index, where 1 represents a valid point and 0 represents an invalid point to be filtered.
Next, we use the index to compute weighted sum of the altitude embedding, and then calculate average altitude value by dividing the number of valid points in the pillar. 
This reduces the dimensionality of the altitude information.
Subsequently, we normalize the average altitude of each pillar to the range of $[0, 1],$ generating 2D altitude encoding $\mathcal{A}^{i} \in \mathbb{R}^{X \times Y}$, which represents the altitude information for each 2D grid.

To retain further 3D context, we compute weighted average of the 3D features along the Z-axis and concatenate it with $\mathcal{A}^{i}$. 
The resulting representation is then passed through a 2D convolution layer for additional compression, producing the final compressed 2D feature ${F}^{i}_{\mathrm{bev}}$. 
Comprehensive compression procedure can be formalized as
\begin{equation} 
{F}^{i}_{\mathrm{bev}} = \Psi\left(\frac{1}{|Z|}\sum_{z} \left( \mathbf{M}^{i}_{z} \odot {F}^{i}_{\mathrm{occ}} \right) + \mathcal{A}^{i}(x, y)\right),
\end{equation}
where
$\sum_{z} (\cdot)$ denotes the summation over the $z$ axis to reduce the 3D feature into 2D representation.
$\mathbf{M}^{i}_{z}$ represents a binary mask that identifies valid points in the feature map based on logistic function and threshold.
$\odot$ denotes element-wise multiplication between the binary mask $\mathbf{M}^{i}_{z}$ and the 3D feature ${F}_{\mathrm{occ}}$.
$\mathcal{A}^{i}(x, y)$ is the altitude encoding for each spatial position $(x, y)$, retaining height information.
$\Psi(\cdot)$ represents the 2D convolution layer for further feature compression after concatenation.
${F}^{i}_{\mathrm{bev}}$ maintains key spatial information while significantly reducing dimensionality, making it more bandwidth-efficient.

\subsection{Dual-Mask Perceptual Guidance \label{sec:dmpg}} 
With the integration of AAR, we obtain altitude-aware planar features. 
In contrast to autonomous driving, where onboard cameras have minimal overlap, UAV mission scenarios involve significant overlap in observation areas between UAVs.
Additionally, UAV observations vary in quality due to occlusions or edge distortions, leading to regions of different observational quality.
To improve data efficiency, we propose Dual-Mask Perceptual Guidance, shown in Fig.~\ref{fig:detail}(b), with a generation processes for a request mask $\mathbf{M}^{i}_\mathrm{req}$ and a support mask $\mathbf{M}^{i}_\mathrm{sup}$. 
$\mathbf{M}^{i}_\mathrm{req}$ identifies areas of poor observation, while $\mathbf{M}^{i}_\mathrm{sup}$ selects high-quality regions for data transmission. 
Each grid in the BEV feature map ${F}^{i}_{\mathrm{bev}}$ is assigned a quality score based on both distance and feature gradient. 
Generation of $\mathbf{M}^{i}_\mathrm{sup}$ can be formalized as
\begin{equation} 
    \mathbf{M}^{i}_{\mathrm{sup}}(x, y) = \begin{cases} 1 & \text{if } \alpha \cdot \frac{h}{\sqrt{h^2 + d^2}} + \beta \cdot \frac{|G(x, y)|}{\epsilon} > \xi \\ 0 & \text{otherwise} \end{cases},
\end{equation}
where $\alpha$ and $\beta$ are weighting coefficients, $h$ represents the UAV's altitude, $d$ is the horizontal distance to the grid $(x, y)$, $|G(x, y)|$ is the gradient magnitude, $\epsilon$ limits gradient complexity, and $\xi$ represents the quality score threshold.
High thresholds restrict data but risk insufficient information; low thresholds transmit more data but can dilute useful features with noise.
The impact of quality score threshold is discussed in the ablation study.
The underperforming regions in $\mathbf{M}^{i}_\mathrm{req}$ are defined as the inverse of $\mathbf{M}^{i}_\mathrm{sup}$. 

In each communication round, ego UAV broadcasts request mask $\mathbf{M}^{i}_\mathrm{req}$ to request high-quality data from neighboring UAVs. 
Upon receiving $\mathbf{M}^{i}_\mathrm{req}$, collaborative UAVs project their support masks $\mathbf{M}^{\left\{1,2,...k \right\} }_\mathrm{sup}$ BEV features and ${F}^{\left\{1,2,...k \right\} }_{\mathrm{bev}}$ into the ego UAV’s BEV space.
They then apply both $\mathbf{M}^{i}_\mathrm{req}$ and the projected $\mathbf{M}^{\left\{1,2,...k \right\} }_\mathrm{sup}$ to extract the features for transmission ${F}_{\mathrm{bev}}^{\mathrm{trans}}$, denoted as
\begin{equation}
    {F}_{\mathrm{bev}}^{\mathrm{trans}} =  \left(\mathbf{M}^{i}_\mathrm{req} \cap \tau \mathbf{M}^{\left\{1,2,...k \right\}}_\mathrm{sup}  \right) \odot \tau {F}^{\left\{1,2,...k \right\} }_{\mathrm{bev}},
\end{equation}
where $\tau$ denotes the transformation to the ego UAV's reference frame. 
This ensures that ego UAV receives only essential features, minimizing redundant data transmission.

\subsection{Cross-Agent Feature Integration \label{sec:cafi}}
We propose Cross-Agent Feature Integration (CAFI) to integrate 2D transmitted feature ${F}_{\mathrm{bev}}^{\mathrm{trans}}$ from connected UAVs with 3D ${F}^{\left\{1,2,...k \right\}}_{\mathrm{occ}}$ of ego UAV.
CAFI restores geometric and semantic information through upsampling and fusion, then output semantic occupancy via a task-specific head.

\noindent\textbf{Upsampling.} 
We upsample $F_{{bev}}^{\text{trans}}$ to improve feature resolution, followed by a 3D convolution to extend it into the 3D space, generating volumetric features. 
The resulting feature implicitly retains altitude information, allowing the feature to effectively capture detailed semantic variations in the 3D environment, particularly along the altitude dimension.

\noindent\textbf{Feature Fusion.} 
The upsampled $ F_{\mathrm{bev}}^{\mathrm{trans}}$ is then concatenated with ego UAV's 3D feature  $F_{\mathrm{occ}} $ along the channel dimension. 
The concatenated features are then processed by a residual 3D convolutional module, yielding fused occupancy feature $F_{\mathrm{occ}}^{\mathrm{fused}}$.
To retain sufficient spatial granularity, we use 3D deconvolutions to refine the fused resolution, ensuring high-quality feature representation.

\noindent\textbf{Task Output.} For fine-grained semantic scene prediction, we utilize a Multilayer Perceptron (MLP) as the task head for semantic segmentation of the fused 3D features, ultimately producing the final collaborative prediction.
Our training approach involves two loss functions.
One is semantic segmentation loss $\mathbf{L}_\text{seg}$, which leverages focal loss to mitigate class imbalance.
The second is communication constraint loss $ \mathbf{L}_\text{com}$, which incorporates $L1$ regularization to minimize data transmission overhead.
The final optimization objective function is given by: $ L = \mathbf{L}_\text{seg} + \lambda \mathbf{L}_\text{com}$.

\section{Experiment}

\definecolor{nothers}{RGB}{139, 137, 137}
\definecolor{nground}{RGB}{200, 200, 200}
\definecolor{nbuilding}{RGB}{128, 128, 255}
\definecolor{nvegetation}{RGB}{0, 175, 0}
\definecolor{nvehicle}{RGB}{255, 150, 0}
\definecolor{nroad}{RGB}{213, 213, 213}
\definecolor{nblue}{RGB}{0, 0, 255}
\definecolor{ngreen}{RGB}{20, 240, 30}
\begin{table*}[ht]
    \footnotesize
    \setlength{\tabcolsep}{0.02\linewidth}
    \begin{center}

    \setlength{\abovecaptionskip}{0pt} 
    \setlength{\belowcaptionskip}{-10pt} 
    \begin{tabular}{c|c|c|c|c|c|c|c|c}
        \toprule
        Dataset & Type & Method & \makecell[c]{Image\\Size} & Co. & \makecell[c]{Range\\(m²)} & \makecell[c]{Height\\(m)} & \makecell[c]{CV(MB)\\$\downarrow$} & \makecell[c]{mIoU\\$\uparrow$} \\
        \midrule
        \multirow{5}{*}{Air-to-Pred-Occ}  
        & \multirow{5}{*}{Simulated} 
        & BEVDet$\dag$   & 1600×900 & ×          & 100×100   & 50 & - & 7.46  \\
        & & PanoOcc        & 1600×900 & ×          & 100×100   & 50 & - & 40.82 \\
        & & BEVDet$\ddag$  & 1600×900 & \checkmark & 100×100 & 50 & 17.50 & 12.29 \\
        & & PanoOcc$\ddag$ & 1600×900 & \checkmark & 100×100 & 50 & 19.14 & 41.96 \\
        & & MCOP (Ours) & 1600×900 & \checkmark & 100×100 & 50 & \textbf{0.23} & \textbf{46.41} \\
        \midrule
        
        \multirow{5}{*}{UAV3D-Occ} 
        & \multirow{5}{*}{Simulated} 
        & BEVDet$\dag$ & 800×450 & × & 112×112 & 60 & - & 8.21 \\
        & & PanoOcc & 800×450 & × & 112×112 & 60 & - & 43.48 \\
        & & BEVDet$\ddag$ & 800×450 & \checkmark & 112×112 & 60 & 17.50 & 12.09 \\
        & & PanoOcc$\ddag$ & 800×450 & \checkmark & 112×112 & 60 & 19.14 & 44.73 \\
        & & MCOP (Ours) & 800×450 & \checkmark & 112×112 & 60 & \textbf{0.23} & \textbf{47.89} \\
        \midrule
        
        \multirow{5}{*}{GauUScene-Occ} 
        & \multirow{5}{*}{Real} 
        & BEVDet$\dag$ & 5472×3648 & × & 500×500 & 150 & - & 7.27 \\
        & & PanoOcc & 5472×3648 & × & 500×500 & 150 & - & 40.43 \\
        & & BEVDet$\ddag$ & 5472×3648 & \checkmark & 500×500 & 150 & 17.50 & 11.18 \\
        & & PanoOcc$\ddag$ & 5472×3648 & \checkmark & 500×500 & 150 & 19.14 & 42.69 \\
        & & MCOP (Ours) & 5472×3648 & \checkmark & 500×500 & 150 & \textbf{0.23} & \textbf{42.92} \\
        \bottomrule
    \end{tabular}
    \end{center}
    \vspace{-15pt} 
    \caption{\textbf{Experimental results of different methods on various datasets.} Co. represents whether collaborative perception is applied. Range represents the observation area of UAV, Height refers to the UAV's flight altitude, and CV denotes the communication volume, which is the data transmission cost per communication instance, measured in MB. $\dag$ indicates that BEV features are converted into occupancy features using the FlashOcc~\cite{flashocc} method for a fair comparison. $\ddag$ denotes the addition of a collaboration module following the Where2comm~\cite{where2comm} method. Our method achieves the highest mIoU and the lowest communication volume.}
    \label{tab:main_exp}
    \vspace{-5pt}

\end{table*}

\subsection{Datasets}
Due to the lack of a suitable dataset for collaborative UAV occupancy prediction, we incorporate semantic occupancy annotations into three datasets: Air-Co-Pred-~\cite{Wang2024DronesHD}, UAV3D~\cite{uav3d2024}, and GauUScene~\cite{xiong2024gauuscene}. 
\textbf{Air-Co-Pred}~\cite{Wang2024DronesHD} is a Carla-based~\cite{dosovitskiy2017carla} virtual dataset feature four UAVs monitoring a 100m×100m intersection at a 50m altitude. It has 32,000 synchronized images (1600×900) split into 170 training and 30 validation scenes.
\textbf{UAV3D}~\cite{uav3d2024} is a synthetic dataset created with Carla~\cite{dosovitskiy2017carla} and AirSim~\cite{shah2018airsim}, covering both urban and suburban environments.
Five UAVs fly at 60m altitude, producing 700 training, 150 validation, and 150 test scenes (800×450). 
We use one town from each virtual dataset for experiments.
\textbf{GauUScene}~\cite{xiong2024gauuscene} is a real-world dataset designed for 3D reconstruction, featuring multiple 1km²-scale scenes with UAV-captured RGB images (5472×3648 resolution), corresponding poses, and point clouds. 
We use one subset covering 0.908km² (“Russian Building” scene) with UAV flights up to 150m altitude. 
Since GauUScene~\cite{xiong2024gauuscene} is not intended for collaborative perception, we treat its four UAV trajectories in this subset as a single four-UAV cluster to enable cooperative sensing.

\noindent\textbf{Occupancy Annotation.}
The original Air-Co-Pred~\cite{Wang2024DronesHD} and UAV3D~\cite{uav3d2024} datasets only provide 2D and 3D bounding box annotations for vehicles.
To facilitate occupancy prediction, we generate additional occupancy annotations for these datasets. 
Specifically, for Air-Co-Pred-Occ and UAV3D-Occ, we first export mesh maps from CARLA~\cite{dosovitskiy2017carla}, and then annotate semantic labels using a 3D point cloud annotation tool. 
For GauUScene-Occ~\cite{xiong2024gauuscene}, we follow the methodology in Occ3D~\cite{occ3d} by reconstructing meshes and assigning corresponding semantic labels. 
All annotations are further refined via ray-casting to realistically simulate occlusions from a single UAV's viewpoint (e.g., objects obscured by walls remain invisible).
Final occupancy annotations include seven semantic categories: \textit{free}, \textit{others}, \textit{ground}, \textit{building}, \textit{vegetation}, \textit{vehicle}, and \textit{urban road}. 
Here, \textit{free} represents unoccupied space as the complement of occupied regions, while \textit{others} denotes objects without specific semantic labels.

\noindent\textbf{Evaluation metrics.} 
Following the evaluation approach for semantic occupancy prediction in autonomous driving, we use Intersection over Union (IoU) as the evaluation metric. 
This involves computing IoU for each class and the mean IoU (mIoU) across all classes. 

\subsection{Experiment Settings}
\noindent\textbf{Implementation Details.} 
For voxelization, Air-Co-Pred-Occ~\cite{Wang2024DronesHD} and UAV3D-Occ~\cite{uav3d2024} use a $0.4m^3$ voxel size, while GauUScene-Occ~\cite{xiong2024gauuscene} adopts a coarser voxel size of $2m^3$ due to the larger observation space.
Air-to-Pred-Occ~\cite{Wang2024DronesHD} and GauUScene-Occ~\cite{xiong2024gauuscene} each  involve four UAVs, while UAV3D-Occ~\cite{uav3d2024} uses five. 
Each UAV covers an observation range with overlapping regions for enhanced perception. 
We employ a ResNet101-DCN~\cite{dai2017deformable} backbone and FPN~\cite{lin2017feature} at four scales (1/8, 1/16, 1/32, 1/64). 
Dual-Mask Perceptual Guidance compresses features via two 2D convolutions with a 0.8 quality threshold. 
Cross-Agent Feature Integration uses hierarchical 3D convolutions for upsampling, and our segmentation head applies two MLP layers (hidden size 128) with softplus~\cite{zheng2015improving} activation.

\noindent\textbf{Training.} 
Training is conducted on eight NVIDIA A6000 GPUs, with a batch size of 1 per GPU. 
We train for 24 epochs using the Adam optimizer with an initial learning rate of $2\times 10^{-4}$ and applying a cosine annealing schedule. 
Data augmentation includes random scaling, cropping, color distortion, and Gridmask~\cite{chen2020gridmask}. 
Each voxel receives a single label from the pre-generated occupancy ground truth.


\begin{figure}[b]
    \vspace{-10pt}
  \centering
    \includegraphics[width=1.0\linewidth]{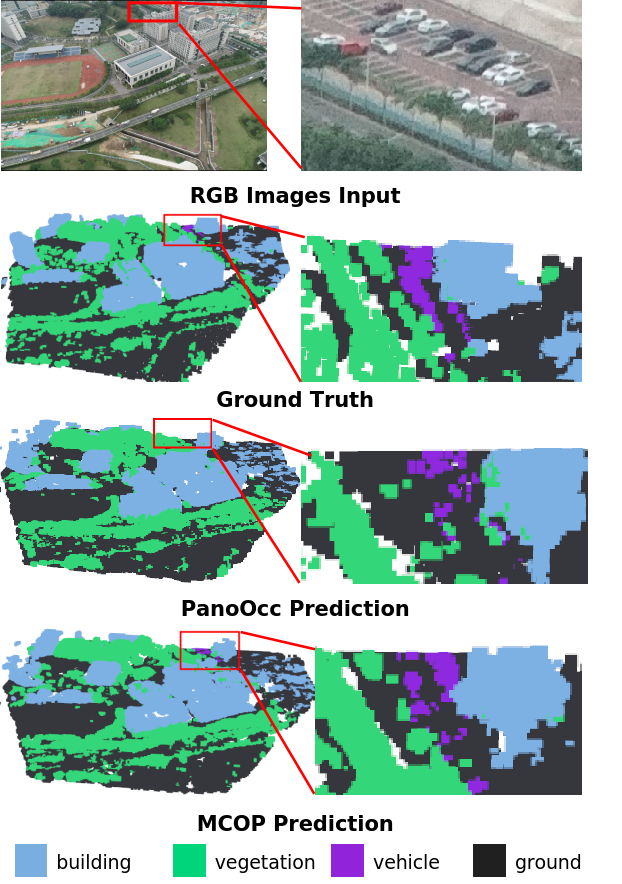}
    \vspace{-20pt}
    \caption{
    \textbf{Visualization results on GauUScene-Occ.} MCOP achieves better perception for distant and occluded objects.}
    
  \label{fig:res}
\end{figure}

\subsection{Comparative Analysis}
Since no occupancy-based approaches exist for UAV perception, we select two representative methods from the autonomous driving domain, BEVDet~\cite{bevdet} and PanoOcc~\cite{panoOcc}, and adapt them for our extended datasets. 
Because BEVDet~\cite{bevdet} only produces 2D BEV predictions, we employ FlashOcc~\cite{flashocc} to convert BEV features into 3D occupancy features, thereby allowing a consistent comparison with PanoOcc~\cite{panoOcc}, which directly outputs occupancy predictions.
As shown in Table~\ref{tab:main_exp}, our experiments demonstrate that adopting occupancy features is advantageous for capturing both geometric and semantic information in 3D environments. 
Moreover, multi-UAV collaboration yields further gains in perception accuracy due to the widened coverage and shared information.

To highlight the advantages of our collaborative strategy, we implement collaborative approaches by integrating the Where2comm~\cite{where2comm} module into BEVDet~\cite{bevdet} and PanoOcc~\cite{panoOcc} for multi-UAV occupancy prediction. 
Experimental results demonstrate that our collaborative strategy achieves superior performance with significantly reduced communication overhead. 
Specifically, our method requires only 0.23 MB per transmission, while BEVDet and PanoOcc demand 17.50 MB and 19.14 MB, respectively. 
Furthermore, on the Air-to-Pred-Occ~\cite{Wang2024DronesHD} dataset, our approach surpasses BEVDet by 35.12 mIoU points and PanoOcc by 4.45 points. 
These results confirm that compressing 3D occupancy features into altitude-aware 2D representations for transmission is more effective and efficient than direct 2D BEV transmission, balancing perceptual accuracy and communication efficiency effectively.
Figure~\ref{fig:res} shows sample results on GauUScene-Occ.

\begin{table}[t]
    \small
    \begin{center}

     \setlength{\tabcolsep}{0.01\linewidth}
     
    \begin{tabular}{c|c|c|c|c}
        \toprule
        Compress & Feature & \multirow{2}{*}{CV(MB)$\downarrow$} & \multirow{2}{*}{mIoU$\uparrow$} & Accuracy \\
        Method & Dimensions &  &  & Loss$\downarrow$\\
        \midrule
        -         & 3D    & 76.56   & 47.17 & -  \\
        \midrule
        Avg.         & 2D  & 1.19    & 30.24 & 16.93\%         \\
        Conv.        & 2D & 1.19    & 41.52 & 5.64\%             \\
        \midrule
        AAR         & 2D  & 1.19    & 46.42 & \textbf{0.75\%}   \\
        \bottomrule
	\end{tabular}
    \end{center}
    \vspace{-10pt}  
    \caption{\textbf{Different Feature Compression Strategies} Avg. represents the weighted average of features along the z-axis. Conv. refers to using a 3D convolution to convert 3D features into 2D features. Accuracy Loss represents the mIoU reduction ratio caused by compressing the features from 3D to 2D.}
    \label{tab:abl_aar}
    \vspace{-5pt}
    
    \end{table}

\begin{table}[t]
    \small
    \begin{center}

     \setlength{\tabcolsep}{0.01\linewidth}
    \begin{tabular}{c|c|c|c|c}
        \toprule
        Connected & Quality & \multirow{2}{*}{CV(MB)$\downarrow$} & \multirow{2}{*}{mIoU$\uparrow$} & Accuracy \\
        Strategy & Threshold &  &  & Loss$\downarrow$\\
        \midrule
        Fully connected & -   & 1.19  & 46.42 & -  \\
        \midrule
        Partially Connected & 0.6 & 0.47  & 45.78 & 0.64\% \\
        Partially Connected & 0.7 & 0.35  & 45.81 & 0.60\%   \\
        Partially Connected & 0.8 & 0.23  & 46.41 & \textbf{0.01\%}   \\
        Partially Connected & 0.9 & 0.11  & 44.97 & 1.45\%  \\
        \bottomrule
	\end{tabular}
    \end{center}
    \vspace{-10pt}
    \caption{\textbf{Different Quality Score Threshold in DMPG.} All the above methods utilize the AAR module, and both Fully Connected and Partially Connected are based on 2D features.}
    \label{tab:abl_dmpg}
    \vspace{-15pt}

    \end{table}

\subsection{Ablation study}
We assess the effectiveness of our modules by removing components under the same settings. 
Since our main experiments already show the advantage of generating and compressing 3D occupancy features over transmitting 2D BEV features, we focus on Altitude-Aware Reduction (AAR) and Dual-Mask Perceptual Guidance (DMPG), and also vary the number of UAVs.

\noindent{\textbf{Effectiveness of Altitude-Aware Reduction.}}
In Table~\ref{tab:abl_aar}, we compare our proposed Altitude-Aware Reduction with two simpler compression methods: a weighted average along the z-axis and 3D convolution for compressing 3D features into 2D. 
All comparisons are conducted without the DMPG module. 
The comparison is performed by evaluating the prediction performance of the compressed features against the uncompressed ones.
The uncompressed features retain the full 3D occupancy representation, resulting in a per-transmission communication cost of 76.56 MB.
Our method reduces accuracy by only $0.75\%$ at the same compression ratio, which is significantly better than the other two methods.
This demonstrates that the introduced altitude encoding effectively preserves altitude information during compression, enabling the restoration of more comprehensive geometric and semantic details in feature fusion.

\noindent{\textbf{Effectiveness of Dual-Mask Perceptual Guidance.}}
We also test different quality score thresholds, which govern whether UAVs transmit features based on perceived quality.
A higher quality score threshold filters out regions of interest, thereby reducing the amount of data transmitted but risking insufficient information for accurate occupancy predictions.
Conversely, a lower quality score threshold, while resulting in more regions being transmitted, can paradoxically lead to a decline in perception quality due to the inclusion of excessive irrelevant information.
To evaluate the effect of different quality score threshold settings, we compare the mIoU accuracy loss between Partially Connected and Fully Connected modes. 
In this context, Partially Connected refers to transmitting only a portion of the features, while Fully Connected involves transmitting the complete set of features.
Our baseline for comparison is the transmission of fully compressed features after applying the AAR module. 
We then assess various quality score threshold settings by comparing the mIoU accuracy loss between Partially Connected and Fully Connected modes. 
Table~\ref{tab:abl_dmpg} shows that a 0.8 threshold achieves the best balance.

\begin{table}[t]
\centering

\small
\begin{tabular}{@{}c|ccccc@{}}
\toprule
\multirow{2}{*}{\textbf{Dataset}}  & \multicolumn{4}{c}{\textbf{UAV Nums (mIoU $\uparrow$)}} \\
\cmidrule(lr){2-6}
 & 1 & 2 & 3 & 4 & 5 \\
\midrule
UAV3D & 43.48 & 46.91 & 47.04 & 47.33 & 47.89 \\
Air-Co-Pred & 40.82 & 45.89 & 46.23 & 46.41 & \textendash \\
GauUScene  & 40.43 & 42.11 & 42.47 & 42.92 & \textendash \\
\bottomrule
\end{tabular}
\vspace{-5pt}
\caption{\textbf{Impact of UAV quantity changes.} UAV num = 1 indicates no collaborative perception.}
\label{tab:abl_uav_num}
\vspace{-15pt}

\end{table}

\noindent{\textbf{Impact of UAV quantity changes.}}
Table~\ref{tab:abl_uav_num} indicates that, under the same scenario, reducing the number of UAVs results in less than a 1\% drop in perception accuracy, which remains higher than without collaborative perception. This demonstrates the robustness of our method to variations in UAV numbers.
\section{Conclusion}
This work proposes a multi-UAV collaborative occupancy prediction framework that addresses key limitations of existing BEV-based methods. 
Our framework effectively captures comprehensive geometric and semantic scene information while significantly reducing communication overhead. 
Extensive experiments on both virtual and real-world datasets demonstrate that our approach outperforms non-collaborative and alternative collaborative strategies, while requiring notably lower bandwidth.

\section*{Acknowledgements}

This work was supported in part by the National Natural Science Foundation of China (No. U21B2042, No. 62320106010). The authors would also like to thank the Key Laboratory of Target Cognition and Application Technology (TCAT), Aerospace Information Research Institute, Chinese Academy of Sciences, Beijing, 100190, China, for their valuable support. In particular, we are grateful to Zhirui Wang and Peirui Cheng for their insightful discussions and assistance throughout the course of this research.

{
    \small
    \bibliographystyle{ieeenat_fullname}
    \bibliography{main}
}
\end{document}